\documentclass[10pt,twocolumn,letterpaper]{article}

\usepackage{cvpr}              % To produce the CAMERA-READY version
\usepackage{graphicx}
\usepackage{amsmath}
\usepackage{amssymb}
\usepackage{booktabs}
\usepackage{times}
\usepackage{epsfig}
\usepackage{multirow}
\usepackage{algorithm}
\usepackage{algorithmic}
\usepackage{bbm}
\usepackage{lipsum}
\usepackage[utf8]{inputenc}

\usepackage[pagebackref,breaklinks,colorlinks]{hyperref}

\newcommand{\x}{\boldsymbol{x}}

\newcommand{\f}{\boldsymbol{f}}

\newcommand{\R}{\mathbb{R}}

\usepackage[capitalize]{cleveref}
\Crefname{table}{Table}{Tables}
\crefname{table}{Tab.}{Tabs.}
\crefname{section}{§}{§§}
\Crefname{section}{§}{§§}

\begin{document}

\title{Exploiting Both Domain-specific and Invariant Knowledge via a Win-win Transformer for Unsupervised Domain Adaptation }

% \author{Wenxuan Ma,\textsuperscript{$\ast$} Jinming Zhang,\textsuperscript{$\ast$} Shuang Li,\textsuperscript{$\dagger$} \\
% {\tt\small \{wenxuanma, jinming-zhang, shuangli\}@bit.edu.cn,}
% \and
% Chi Harold Liu, Yulin Wang, Wei Li\\
% % First line of institution2 address\\
% {\tt\small liuchi02@gmail.com, wang-yl19@mails.tsinghua.edu.cn, liweimcc@gmail.com}
% }
\author{
 Wenxuan Ma\textsuperscript{1$\ast$} \space\space
 JinMing Zhang\textsuperscript{1$\ast$} \space
 Shuang Li\textsuperscript{1$\dagger$} \space \space
 Chi Harold Liu\textsuperscript{1} \space\space\space
 Yulin Wang\textsuperscript{2} \space\space
 Wei Li\textsuperscript{3} \vspace{.3em}\\
 \textsuperscript{1}Beijing Institute of Technology \quad \textsuperscript{2}Tsinghua University \quad \textsuperscript{3}Inceptio Tech.\\
 \vspace{-.3em}
 {\tt\small \{wenxuanma, jinming-zhang, shuangli\}@bit.edu.cn}\\
 {\tt\small liuchi02@gmail.com} \space \space {\tt\small wang-yl19@mails.tsinghua.edu.cn} \space \space {\tt\small liweimcc@gmail.com} 
 \vspace{-.5em}
}

\maketitle
\newcommand\blfootnote[1]{%
  \begingroup
  \renewcommand\thefootnote{}\footnote{#1}%
  \addtocounter{footnote}{-1}%
  \endgroup
}

\begin{abstract}
Unsupervised Domain Adaptation (UDA) aims to transfer knowledge from a labeled source domain to an unlabeled target domain. Most existing UDA approaches enable knowledge transfer via learning domain-invariant representation and sharing one classifier across two domains. However, ignoring the domain-specific information that are related to the task, and forcing a unified classifier to fit both domains will limit the feature expressiveness in each domain. In this paper, by observing that the Transformer architecture with comparable parameters can generate more transferable representations than CNN counterparts, we propose a \textbf{Win}-Win \textbf{TR}ansformer framework (WinTR) that separately explores the domain-specific knowledge for each domain and meanwhile interchanges cross-domain knowledge. Specifically, we learn two different mappings using two individual classification tokens in the Transformer, and design for each one a domain-specific classifier. The cross-domain knowledge is transferred via source guided label refinement and single-sided feature alignment with respect to source or target, which keeps the integrity of domain-specific information. Extensive experiments on three benchmark datasets show that our method outperforms the state-of-the-art UDA methods, validating the effectiveness of exploiting both domain-specific and invariant information for both domains in UDA. 
\blfootnote{$\ast$ Equal contribution}
\blfootnote{$\dagger$ Corresponding author}
\end{abstract}

\section{Introduction}\label{sec:introduction}

It is well known that deep neural networks are data-hungry~\cite{vgg,resnet} and hard to generalize to test data sampled from a different distribution. Therefore, despite that they have achieved remarkable success on multiple computer vision tasks such as image classification~\cite{resnet}, object detection~\cite{fasterRCNN} and semantic segmentation~\cite{deeplabv3}, their real-world utility is often limited. To address these practical issues, unsupervised domain adaptation (UDA) is proposed, aiming to transfer knowledge from a labeled source domain to another unlabeled target domain with the presence of \textit{domain shift}~\cite{survey,ben2007analysis} in data distribution.

\begin{figure}[tb]
    \centering
    \includegraphics[width=0.48\textwidth]{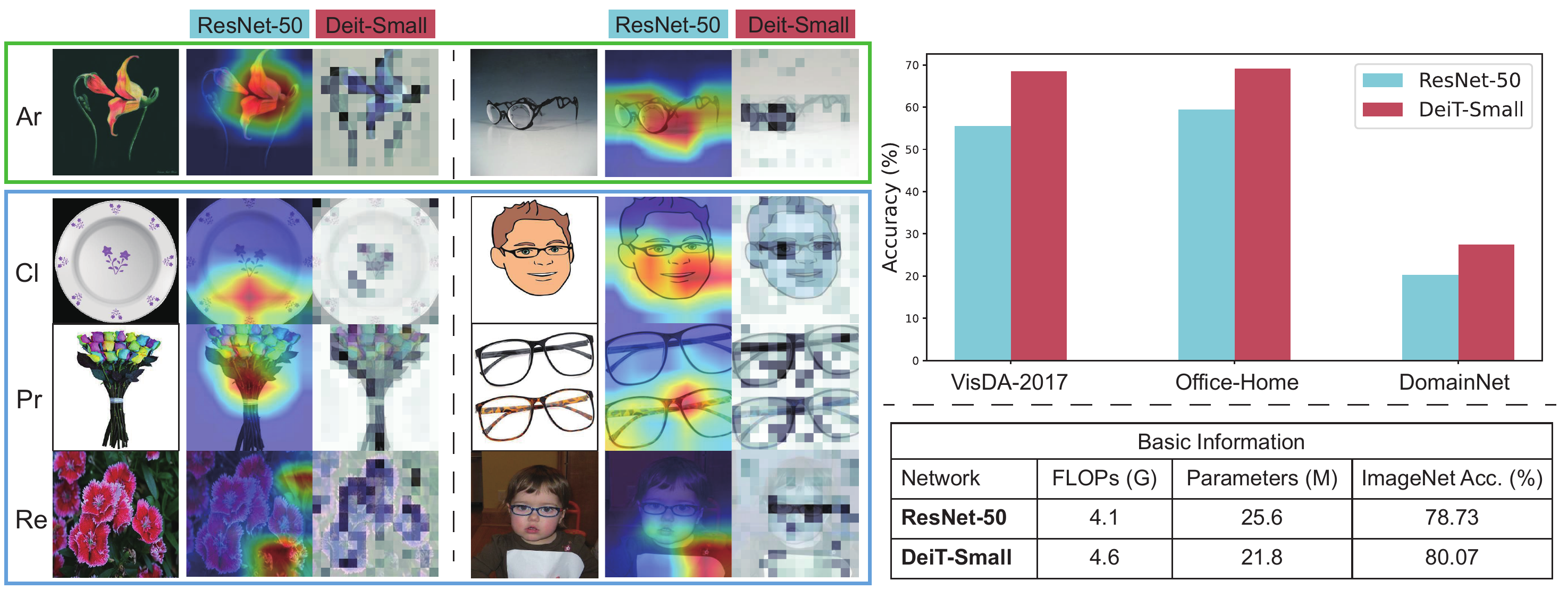}
    \caption{Qualitative (left) and quantitative (right) comparisons of the transferability between CNN and vision Transformer. The left half shows the attention visualization on all four domains in Office-Home, where both models are pretrained on domain \textbf{Ar}. CNN fails to focus on the main objects in the target domains, yet vision Transformer does. The right half shows the average target accuracy on three benchmarks with respective source pretrained model. Despite both architectures with comparable size and FLOPs achieve similar performances on ImageNet, their fine-tuned source models show significant transferability difference.}
    \label{Fig_illstration}
\end{figure}

Over the past decade, researchers have made great effort to improve the performance on UDA. Following the guidance of the theory~\cite{ben2010theory}, most works adopt a similar paradigm that \textit{learns to extract domain-invariant representations from the encoder and reuse the source classifier on top of that}. In vision tasks, the basic architecture of the encoder is usually a convolution-based neural network such as VGG~\cite{vgg} and ResNet~\cite{resnet}. Domain-invariant representations are learned via the domain alignment process, which is either explicit (e.g. minimizing certain statistical metrics~\cite{DAN,JAN,CORAL} or confusing the domain discriminator~\cite{DANN,CDAN}) or implicit (like self-training~\cite{CBST,SPCAN,atdoc} or target-oriented data augmentation~\cite{TSA}). However, due to lacking of true label supervision in the target domain, the class-discriminative information is easily lost during domain alignment~\cite{BSP}, and the source classifier still inevitably biases to the source domain~\cite{atdoc}, damaging the effectiveness of knowledge transfer. We generally interpret these problems as a dilemma between exploring the knowledge from its own domain and acquiring the knowledge transferred from the other domain. Such a dilemma leads to the following question: can one develop a win-win solution for in-domain knowledge preservation and cross-domain knowledge transfer in UDA?

Solution to the question above may come with vision Transformers. Recently, the rise of Transformer in vision tasks shows the potential of pure attention-based models. One prominent difference of these models is that they abandon the strong inductive bias hard-coded in the convolutional neural networks (CNNs)~\cite{convit,conv-v-trans}. The inductive bias in the form of locality and spacial invariance makes convolution-based models more sample-efficient and parameter-efficient and achieve strong performances~\cite{convit}, yet it has certain limitations in learning the cross-domain transferable representation. For instance, the local convolution operation intermixes the task-related content information with task-irrelevant style information such as the background color, making the high-level representation captured by CNN affected by task-irrelevant styles~\cite{GDCAN}. Transformers, on the other hand, do not confined to these restrictions and thus learns more transferable representations~\cite{conv-v-trans}.
% As shown in Fig.~\ref{Fig_illstration}, we can limitations imposed by the encoded inductive bias makes the representations of CNNs less transferable than those of Transformers, in the way that CNN models pretrained in source domain pay attention to the wrong place in the target images more easily, and their performances on target domain are far behind than their vision Transformer counterparts which have tantamount network parameters and classification accuracy on ImageNet. 
As shown in Fig.~\ref{Fig_illstration}, we empirically observe that CNN models pretrained on source domain pay attention to the wrong place in the target images more easily, especially under large domain gap (e.g., Ar$\rightarrow$Cl in Office-Home), and their performances on target domain are far behind their vision Transformer counterparts which have tantamount network parameters and classification accuracy on ImageNet~\cite{imagenet}. 

Another intriguing property of vision Transformer lies in its flexibility of appending learnable classification tokens that serve for different purposes. Previous UDA methods generally reuse the source classifier for target prediction on top of a shared feature representation. Ideally, the features of the two domains can be perfectly aligned class-wisely and the classifier can make accurate predictions on target domain. However, due to the difficulty of perfect alignment in practice, such paradigm raises a competition. 

On the feature level, the alignment encourages cross-domain knowledge sharing and suppresses domain-specific information. Yet on the classifier level, the source classifier explores source-specific knowledge for minimizing source classification loss.
Some recent works make efforts to balance the two sides~\cite{metaAlign,DWL}, but they are still confined to the paradigm. In this paper, we make an attempt on exploring a new possibility of handling both part of the knowledge.
Inspired by DeiT~\cite{DeiT} that two tokens converge to towards different vectors when trained on different objectives, adopting two tokens that respectively learn the source-oriented and target-oriented representations (or tokens) seems to be a reasonable choice. Accordingly, two classifiers instead of one can be utilized to match different tokens.

Based on these considerations, we propose in this paper a Transformer-based UDA framework dubbed as win-win Transformer to simultaneously exploit domain-specific and invariant knowledge. We adopt two classification tokens [src] and [tgt] that learns different mappings, and build on top of each one a domain-specific classifier. To maximally preserve the domain-specific information, we let the two classifiers learn from labels or pseudo-labels in their own domain respectively, and add attention masks to the self-attention operation to prevent the two classification tokens from intervening each other. To achieve knowledge transfer across source and target domain for mutual enhancement, we propose two cross-domain transfer mechanism: the target pseudo-label refinement mechanism using source structural information and the bi-direction regularization mechanism using the discriminative information.
We empirically choose the weighted K-means clustering~\cite{SHOT} and the supervised contrastive learning~\cite{supCon} as the practical implementation while showing other variants also feasible in~\cref{sec:analysis}.

In general, we highlight our three-fold contributions.

\begin{itemize}
    \item We set up a new transformer-based UDA framework that has two classification tokens for learning different domain-oriented representations and two domain-specific classifiers to fully explore domain-specific knowledge without mutual interference between the two domains.
    \item We propose the source guided label refinement to transfer source structural information to target domain and conduct single-sided feature alignment in both directions through contrastive learning with stop-gradient operation.
    \item Our method achieves the state-of-the-art performance on three benchmark UDA datasets, which might shed light on future researches on addressing transfer learning problems using vision transformers.
\end{itemize}
\section{Related Work}\label{sec:related}
\subsection{Unsupervised Domain Adaptation (UDA)} 
To address the performance degradation issue when deploying models to new environments, UDA is proposed and widely studied~\cite{survey}, which aims to train a model that can adapt from labeled source domain to unlabeled target domain. The mainstream of UDA methods focuses on learning domain-invariant representations and reuserd the source classifier. Statistical metrics such as maximum mean discrepancy (MMD)~\cite{MMD} between source and target domain are proposed as objectives for models to minimize~\cite{DAN,JAN,DRCN}. Other approaches find inspiration from adversarial training~\cite{GAN} and thus capture domain-invariant representations via a min-max game with the domain discriminator. For example, DANN~\cite{DA_bp} introduces a gradient reversal layer to enable simultaneous optimization of the two players. CDAN~\cite{CDAN} brings class conditional information to the discriminator, GVB~\cite{GVB} adds ``gradually vanishing bridges'' to promote alignment and RADA~\cite{RADA} re-energizes the discriminator using dynamic domain labels.

Another line of researches borrows the pseudo-labeling idea from semi-supervised learning~\cite{pseudo-label}, where the reliable model predictions on unlabeled data are chosen as pseudo-labels to assist the model retraining. Most UDA approaches~\cite{CBST,MSTN,ALDA} adopt target pseudo-labels for a better conditional distribution alignment, such as CBST~\cite{CBST} and MSTN~\cite{MSTN}. To obtain less noisy pseudo labels, SHOT~\cite{SHOT} and ATDOC~\cite{atdoc} consider the structural information of target data to refine the original labels. These methods can be regarded as conducting implicit domain alignment by making the source and target features of the same class similar, and also adapt the source classifier to be more target-specific. The combination of the two parts of knowledge makes the pseudo-labeling based UDA achieve promising results. However, these works train both source and target data on the same classifier, which might damage the domain-specific information for both domains.
Our method, on the other hand, individually trains two classifiers using data from the respective domain to maximally preserve domain-specific knowledge.

Recently, some works start to pay more attention to domain-specific information learning\cite{GDCAN,DSBN,DWT-MEC}. To model domain discriminative representation separately, DSBN\cite{DSBN} introduces a domain-specific batch normalization mechanism. GDCAN\cite{GDCAN} designs a domain conditional attention module in convolutional layers to activate distinctly interested channels for each domain, while DWT~\cite{DWT-MEC} makes the domain-specific whitening transform at higher layers. Different from them, our method takes the advantage of Transformer architectures to separately explore source-specific and target-specific knowledge.

\subsection{Vision Transformers}
Motivated by the significant improvements of Transformers on natural language processing (NLP) tasks, researchers begin to apply Transformer architectures to computer vision (CV) as a potential alternative for CNN backbones. ViT~\cite{vit} proposes to apply a Transformer model with image patches as inputs to solve image classification problems. Its competitive results compared to CNN backbones encourages further explorations on attention-based networks in CV. Later, DeiT~\cite{DeiT} introduces a distillation token with more training strategies that enables ViT to train on much smaller datasets. Recent works such as Swin Transformer~\cite{swin}, PVT~\cite{PVT} and CrossViT~\cite{crossvit} improve architecture from different aspects. Furthermore, other researchers apply vision transformers to downstream tasks like semantic segmentation~\cite{SETR,volo}, object detection~\cite{DETR,rethinkingDETR} and multimodal tasks~\cite{unit}. Meanwhile, more detailed comparisons between vision Transformer and CNNs are investigated to show their relative strength and weakness~\cite{convit,conv-v-trans,visformer,DoTransSeeLikeConv}.

Since Transformer has the intrinsic advantages to extract more transferable representations, several works~\cite{TransDA,cdtrans} have been proposed to solve domain adaptation with it. For instance, TransDA~\cite{TransDA} injects an attention module after the CNN-type feature extractor to guide the model attention in source-free DA. CDTrans~\cite{cdtrans} adopts the complete transformer architecture and introduces a cross-attention mechanism between the selected source-target pairs. The outputs of the cross-attention branch are then used to supervise the target self-attention branch. We take a different perspective from these methods by noticing that the classification tokens in vision Transformers is capable of learning different mappings, hence is desirable for preserving domain-specific knowledge. Therefore, we propose to train a transformer with two classification tokens to both exploit in- and cross-domain knowledge for extracting better representations.

\section{Win-win Transformer}
This section we introduce our win-win Transformer framework in detail. Our key notion of win-win is mainly reflected in that the proposed framework allows the learned representation to benefit from both in-domain and cross-domain knowledge, without leading to a zero-sum competition that either losses crucial domain-specific information or leads to domain misalignment.

\begin{figure*}[tb]
  \centering
  \includegraphics[width=0.91\textwidth]{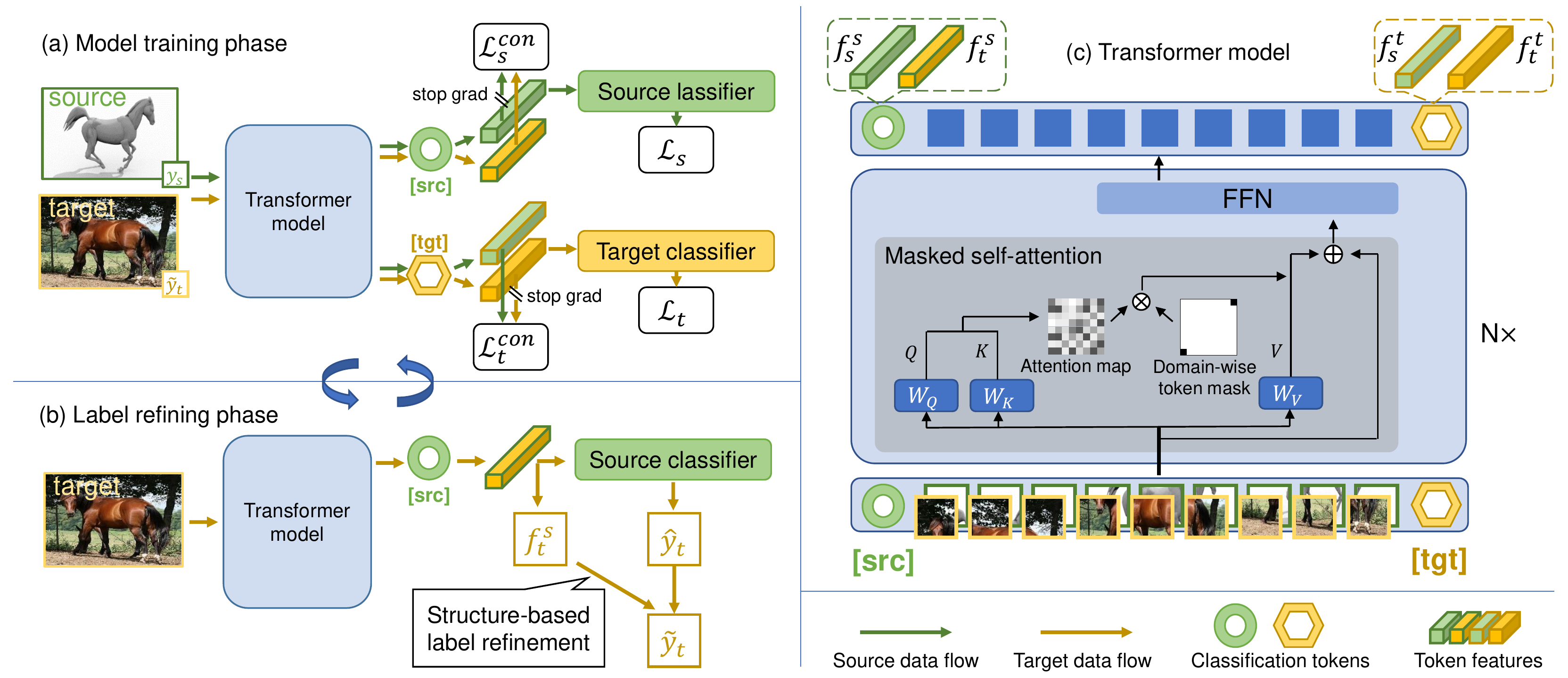}
  \caption{Illustration of the proposed win-win Transformer framework. The model training process contains (a) the model training phase that individually trains two domain-specific classifiers based on domain-oriented representations, with different learning objectives, and (b) the label refining phase that utilizes source structural information to assign better target pseudo-labels. The separately trained domain representations and the proposed single-sided feature alignment enable our framework to achieve a win-win between exploiting in-domain and cross-domain knowledge. Our detailed transformer structure is shown in (c), where two classification tokens [src] and [tgt] are simultaneously sent into the network along with image patches to obtain four different representations (two for each token) $\f^s_s$, $\f^s_t$, $\f^t_s$ and $\f^t_t$. A domain-wise token mask is added on each attention module to keep the two domain-oriented tokens separate. (Best viewed in color.)}
  \label{Fig_method}
  \end{figure*}

\subsection{Preliminaries}
We begin by introducing the notations in UDA and the self-attention mechanism in vision Transformers. Data in UDA are sampled from two different distributions $P_s$ and $P_t$ to create a labeled source domain $D_s=\{\left(\x_{si},y_{si}\right)\}_{i=1}^{n_s}$ and an unlabeled target domain $D_t=\{\x_{tj}\}_{j=1}^{n_t}$, where $n_s$ and $n_t$ denote the number of sampled data. The goal of UDA methods is to utilize these data to train a model that performs well on the target domain. While most prior methods extract feature representations from data using convolution-based models, this work turns to fully attention-based vision Transformer to learn embeddings that can simultaneously benefit from in-domain and cross-domain knowledge.

\textbf{Self-attention mechanism} is at the core of vision Transformers~\cite{vit}. Given a sequence of input features $X\in\R^{N\times D}$, three learnable linear projectors $W_Q$, $W_K$, $W_V$ are applied on the layer-normalized features of it separately to obtain the queries $Q\in\R^{N\times d}$, the keys $K\in\R^{N\times d}$ and the values $V\in\R^{N\times d}$. The query sequence is then matched against the keys using inner products to get the $N\times N$ attention matrix whose elements represent the semantic relevance of the query-key pairs in the corresponding position. The output of the self-attention is the weighted sum of the values, determined according to the attention matrix:
\begin{equation}\label{equ:attn}
  {\rm Attention}(Q,K,V)={\rm softmax}(\frac{Q K^{\rm T}}{\sqrt d})V.
\end{equation}
Self-attention can be viewed as a feature aggregation for each position, using transformed features from other strongly correlated positions. This allows the Transformer to jump out of the inductive bias of locality and incorporate more global information~\cite{DoTransSeeLikeConv}.
The output of the self-attention is combined with the original input through residual connection and further passes through a feed-forward network (FFN), which completes a full self-attention module in vision Transformer.   

\subsection{In-domain Knowledge Exploration with Two Classification Tokens}

The main idea of in-domain knowledge exploration is to enable the model to learn domain-specific information.
% Given source samples and their corresponding ground truth labels, it is straightforward to equip network with basic source classification ability. Thus, following the standard source supervised learning setting, we first use empirical risk minimization for all classifiers on source samples:
Formally, for each input sequence of image features from image patches $X_p = [x_p^1; x_p^2; ... ; x_p^M]\in \R^{M\times D}$, we add two learnable classification tokens [src] and [tgt], namely $X = [x_{[src]};X_p;x_{[tgt]}]\in \R^{N\times D}$ where $N = M+2$. Since we prefer the two tokens to learn different mappings that respectively contains more domain-specific information for source and target, we design a domain-wise token mask to prevent the two token from aggregating each other's feature. 
Specifically, we define an $N \times N$ mask matrix with values at bottom-left and top-right corners equal to negative infinity, which will then become zero after the Softmax function:
\begin{equation}\label{equ:mask}
  M_{ij}=\begin{cases}
    -\infty, & (i,j)=(1,N) \; {\rm or} \; (N,1)\\
    0,& {\rm others}
  \end{cases}.
\end{equation}
Then the mask is applied in each self-attention module, thus substituting the Attention formula in Eq.~\eqref{equ:attn} by the following MaskedAttn.

\begin{equation}\label{equ:masked_attn}
  {\rm MaskedAttn}(Q,K,V)={\rm softmax}(\frac{Q K^{\rm T}}{\sqrt d}+M)V.
\end{equation}

The feature sequence will forward propagate to the last layer of the vision Transformer, and we utilize the two features that corresponds to the classification tokens. The source and target feature generated from the source token (source-oriented) are denoted as $\f^s_{si}=f^s(\x_{si}), \; \f^s_{tj}=f^s(\x_{tj})$, and that from the target token (target-oriented) as $\f^t_{si}=f^t(\x_{si}), \; \f^t_{tj}=f^t(\x_{tj})$. Next, we build a source-specific classifier $g^s$ and a target-specific classifier $g^t$ to separately proceed features $\f^s$ and $\f^t$. Then, the ground truth source label is adopted on $g^s$ to explore the source-specific knowledge: 

\begin{equation}\label{equ:source_loss}
  \mathcal{L}_{s}=\frac{1}{n_s}\sum_{i=1}^{n_s}{\mathcal{E}(\boldsymbol{g}^{s}_{si},{y}_{si})},
\end{equation}
where $\mathcal{E}(\cdot, \cdot)$ denotes the standard cross-entropy (CE) loss, and $\boldsymbol{g}^{s}_{si} = g^s \circ f^s(\x_{si})$ is the logit output.
Similarly, the target-specific classifier and the target-oriented representation are supervised by the target CE loss, using pseudo-labels in the target domain:
\begin{equation}\label{equ:target_loss}
  \mathcal{L}_{t}=\frac{1}{n_t}\sum_{j=1}^{n_t}{\mathcal{E}(\boldsymbol{g}^{t}_{tj},\tilde{y}_{tj})}.
\end{equation}
Here, $\tilde{y}_{tj}$ is the hard pseudo-label for $\x_{tj}$ and $\boldsymbol{g}^{t}_{tj} = g^t \circ f^t(\x_{tj})$. We will further discuss how to obtain it in the next section.
It's worth noting that different from ATDOC~\cite{atdoc} which regards a nonparametric model as the auxiliary target classifier to obtain pseudo-labels for training the main linear classifier, our target classifier is directly trained for target domain classification.

To this end, we replace the traditional shared classifier with two domain-specific classifiers trained on different loss functions. Nevertheless, training a classifier with only target pseudo-labels could be problematic. Hence, it needs information from source domain for correction, and meanwhile an effective way to transfer cross-domain knowledge is demanded. We address these issues by utilizing the ``by-products'' in the process, i.e. the source-viewed target features and target-viewed source features.

\subsection{Bi-direction Cross-domain Knowledge Transfer}
\textbf{Target pseudo-label refinement incorporating source structural information.} Since target pseudo-label is the only information to supervise the target-specific classifier, its noise must be maximally suppressed. We address this by using the structural information, that features in the neighborhood are likely to have the same label, to refine the original pseudo-labels which are based on predictions. The key point here is that we discover that better refinement results are achieved using source-oriented representation instead of the target-oriented one. Our hypothesis is that the source-oriented representation is supervised by clean source labels, which preserves finer structural relationships between feature points. On the other hand, the target-oriented representation is prone to overfit on mislabeled data which damages the structural information. Therefore, by refining target pseudo-labels on source-oriented representations, we remedy the problem through source knowledge. The results of a comparison study are shown in~\cref{sec:analysis}.
% The pretrained model provides a good starting point given the better transferability of the vision Transformer than CNNs, as discussed in section~\ref{sec:introduction}. Namely, we use the one-hot form of the predicted class as the pseudo-label $\hat{y}_t$ at the beginning.

Empirically, we adopt a simplified version of the previously proposed weighted K-means method~\cite{SHOT}, which achieves superior performance especially when the number of categories is large. The class centers are calculated as a weighted sum over all (source-oriented) target features $\f^s_{tj}$, and the refined pseudo-labels are obtained by finding the nearest class center:
\begin{equation}\label{equ:kmeans}
  \begin{aligned}
  \boldsymbol{c}_k =& \frac{\sum_{\x_{tj}\in D_t}\delta_k(\boldsymbol{g}^s_{tj})\f^s_{tj}}{\sum_{\x_{tj}\in D_t}\delta_k(\boldsymbol{g}^s_{tj})}, \\
  \tilde{y}_{tj} &= \arg \min_k d(\boldsymbol{g}^s_{tj}, \boldsymbol{c}_k).
  \end{aligned}
\end{equation}
The function $d(\cdot, \cdot)$ denotes the cosine distance between the two vectors. The label refinement process starts at the very beginning and is repeated with a fixed interval until the network converges.

So far, the source-oriented representation has not yet received any information about the target domain. To enhance its generalization performance on target data without damaging its source-specific information, we propose to conduct feature-level alignment between $\f^s_s$ and $\f^s_t$ with a stop-gradient operation on the former one. This is inspired by works in self-supervised learning~\cite{SimSiam, BYOL} where they use stop-gradient to avoid model collapse. In our case, we expect to keep $\f^s_s$ still and bring $\f^s_t$ closer towards it to transfer target domain knowledge. MMD~\cite{MMD}, MSTN~\cite{MSTN} and contrastive learning with label information~\cite{supCon} are selected as candidates, and we empirically find that contrastive learning prevails. Specifically, given each target sample $\x_t$ with pseudo-label $\tilde{y}_t$, we have positive source samples $\x_s^+$ with the same class label and other negative source samples, and we optimize the following loss

% \begin{equation}\label{equ:shot_y0}
%   \hat{y}_{tj} = \arg \min_k d(\boldsymbol{g}^s_{tj}, \boldsymbol{c}_k), 
% \end{equation}

% \begin{equation}\label{equ:shot_c1}
%   \begin{aligned}
%   \boldsymbol{c}_k^{\prime} =& \frac{\sum_{\x_{tj}\in D_t}\mathbbm{1}(\hat{y}_{tj}=1)\f^s_{tj}}{\sum_{\x_{tj}\in D_t}\mathbbm{1}(\hat{y}_{tj}=1)}, \\
%   \tilde{y}_{tj} &= \arg \min_k d(\boldsymbol{g}^s_{tj}, \boldsymbol{c}_k^{\prime}).
%   \end{aligned}
% \end{equation}

\begin{equation}\label{equ:source_conloss}
  \mathcal{L}_{s}^{con}=-\mathbb{E}_{\x_t}\mathbb{E}_{\x_s^+}\left[{\rm log}\frac{{\rm exp}(f^t(\x_t)\cdot f^t(\x_s^+)/\tau)}{\sum_{\x_s}{\rm exp}(f^t(\x_t)\cdot f^t(\x_s)/\tau)}\right].
\end{equation}

Loss $\mathcal{L}^{con}_s$ is combined with cross-entropy loss $\mathcal{L}_s$ to train the source classifier and [src] token. With the help of target knowledge, the label refinement in Eq.~\eqref{equ:kmeans} obtains less noisy target pseudo-labels that will benefit the target classifier in return. 

\textbf{Target self-training regularization using source discriminative information.}
To further transfer the source knowledge to promote target representation learning, we adopt a symmetric form of Eq.~\eqref{equ:source_conloss} on $\f^t_t$ and $\f^t_s$:

\begin{equation}\label{equ:target_conloss}
  \mathcal{L}_{t}^{con}=-\mathbb{E}_{\x_s}\mathbb{E}_{\x_t^+}\left[{\rm log}\frac{{\rm exp}(f^t(\x_s)\cdot f^t(\x_t^+)/\tau)}{\sum_{\x_t}{\rm exp}(f^t(\x_s)\cdot f^t(\x_t)/\tau)}\right],
\end{equation}
where $\x_t^+$ are positive target samples that have the same (pseudo) labels as $\x_s$ does. Note that this time we stop the gradient from $\f^t_t$.

The overall formulation of our objective function is the sum of four losses with only one trade-off parameter $\lambda$:
\begin{equation}\label{equ:total_loss}
  \mathcal{L} = \mathcal{L}_{s} + \mathcal{L}_{t} + \lambda(\mathcal{L}_{s}^{con}+\mathcal{L}_{t}^{con}).
\end{equation}

\begin{table*}[htbp]
  \centering
  \LARGE
  \caption{Accuracy (\%) on VisDA-2017 for unsupervised domain adaption. ($\sim$ denotes the method having similar amount of parameters to the corresponding backbone.)}
  \label{tab:visda}%
  \vspace{-8pt} 
  \resizebox{0.85\textwidth}{!}{%
    \begin{tabular}{lcccccccccccccc}
    \toprule
    Method & Params (M) & plane & bcycl & bus & car & horse & knife & mcycl & person &plant & sktbrd & train & truck & Avg. \\
 \midrule
 \multicolumn{15}{c}{\textbf{ResNet-101} backbone} \\
 \midrule
 ResNet-101~\cite{resnet} & 44.6 & 55.1 & 53.3 & 61.9 & 59.1 & 80.6 & 17.9 & 79.7 & 31.2 & 81.0 & 26.5 & 73.5 & 8.5 & 52.4 \\
 DANN\cite{DANN} & $\sim$ & 81.9 & 77.7 & 82.8 & 44.3 & 81.2 & 29.5 & 65.1 & 28.6 & 51.9 & 54.6 & 82.8 & 7.8 & 57.4\\
 SWD\cite{SWD} & $\sim$  & 90.8 & 82.5 & 81.7 & 70.5 & 91.7 & 69.5 & 86.3 & 77.5 & 87.4 & 63.6 & 85.6 & 29.2 & 76.4\\
 CDAN+E\cite{CDAN} & $\sim$  & 85.2 & 66.9 & 83.0 & 50.8 & 84.2 & 74.9 & 88.1 & 74.5 & 83.4 & 76.0 & 81.9 &38.0 & 73.9 \\
 BNM\cite{BNM} & $\sim$  & 89.6 & 61.5 & 76.9 & 55.0 & 89.3 & 69.1 & 81.3 & 65.5 & 90.0 & 47.3 & 89.1 & 30.1 &70.4\\
 DWL~\cite{DWL} & $\sim$  & 90.7 & 80.2 & 86.1 & 67.6 & 92.4 & 81.5 & 86.8 & 78.0 & 90.6 & 57.1 & 85.6 & 28.7 & 77.1\\
 MSTN+DSBN\cite{DSBN} & $\sim$  & 94.7 & 86.7 & 76.0 & 72.0 & 95.2 & 75.1 & 87.9 & 81.3 & 91.1 & 68.9 & 88.3 & 45.5 & 80.2\\
 SHOT\cite{SHOT} & $\sim$  & 94.3 & 88.5 & 80.1 & 57.3 & 93.1 & 93.1 & 80.7 & 80.3 & 91.5 & 89.1 & 86.3 & 58.2 & 82.9 \\
 FixBi\cite{fixbi} & $\sim$  & 96.1 & 87.8 & 90.5 & 90.3 & 96.8 & 95.3 & 92.8  &\textbf{88.7} & 97.2 & 94.2&  90.9 & 25.7&  87.2 \\
  \midrule
  \multicolumn{15}{c}{\textbf{DeiT} backbone} \\
  \midrule
  DeiT-S\cite{DeiT}  & 21.8 & 95.7 & 46.3 & 82.9 & 68.7&  83.4 & 57.1 & \textbf{96.3} & 21.8  & 87.5 & 42.8 & 92.8 & 24.7& 66.7\\
  WinTR-S (Ours)& $\sim$ & 97.5 & 87.0 &	91.2 & 88.3 & 97.4 & 95.6 & 94.7 & 79.1 & 96.6 & 93.9 & 93.3 & 51.5 & 88.8 \\ 
  \midrule
  DeiT-B\cite{DeiT} & 86.3 & 97.7 & 48.1 & 86.6 & 61.6 & 78.1 & 63.4 & 94.7 & 10.3 & 87.7 & 47.7 & 94.4 & 35.5 & 67.1\\
  CDTrans-B\cite{cdtrans} & $\sim$  & 97.1 & 90.5 & 82.4 & 77.5 & 96.6 & \textbf{96.1} & 93.6 & 88.6 & \textbf{97.9} & 86.9 & 90.3 & \textbf{62.8} & 88.4 \\
  WinTR-B (Ours) & $\sim$  & \textbf{98.7}  & \textbf{91.2} &  \textbf{93.0} &  \textbf{91.9} &  \textbf{98.1} &  \textbf{96.1} &  94.0 &  72.7 &  97.0 &  \textbf{95.5} &  \textbf{95.3}&   57.9&   \textbf{90.1}\\  
    \bottomrule
  \end{tabular}
  }% 
\end{table*}%

\begin{table*}[htbp]
  \centering
  \Huge
  \caption{Accuracy (\%) on Office-Home for unsupervised domain adaption.}
  \label{tab:home}%
  \vspace{-8pt} 
  \resizebox{0.85\textwidth}{!}{%
    \begin{tabular}{lcccccccccccccc}
    \toprule
    Method & Params (M) & Ar$\rightarrow$Cl & Ar$\rightarrow$Pr & Ar$\rightarrow$Re & Cl$\rightarrow$Ar & Cl$\rightarrow$Pr & Cl$\rightarrow$Re & Pr$\rightarrow$Ar & Pr$\rightarrow$Cl &Pr$\rightarrow$Re & Re$\rightarrow$Ar & Re$\rightarrow$Cl & Re$\rightarrow$Pr & Avg. \\
    \midrule
    \multicolumn{15}{c}{\textbf{ResNet-50} backbone} \\
    \midrule
    ResNet-50~\cite{resnet} & 25.6 & 44.9 & 66.3 & 74.3 & 51.8 &61.9 & 63.6 & 52.4 & 39.1 & 71.2 & 63.8 &45.9 & 77.2 & 59.4 \\
     CDAN+E\cite{CDAN} & $\sim$ & 51.0 & 71.9 & 77.1 & 61.2 & 69.1 & 70.1 & 59.3 & 48.7 & 77.0 & 70.4 & 53.0 & 81.0 & 65.8 \\
     BNM\cite{BNM} & $\sim$  & 56.7 & 77.5 & 81.0 & 67.3 & 76.3 & 77.1 & 65.3 & 55.1 & 82.0 & 73.6 & 57.0 & 84.3 & 71.1\\
     GVB\cite{GVB} & $\sim$  & 57.0 & 74.7 & 79.8 & 64.6 & 74.1 & 74.6 & 65.2 & 55.1 & 81.0 & 74.6 & 59.7 & 84.3 & 70.4\\
     MCC\cite{mcc} & $\sim$  & 56.3 & 77.3 & 80.3 & 67.0 & 77.1 & 77.0 & 66.2 & 55.1 & 81.2 & 73.5 & 57.4 & 84.1 & 71.0\\
     BSP+TSA\cite{TSA} & $\sim$  & 57.6 & 75.8 & 80.7 & 64.3 & 76.3 & 75.1 & 66.7 & 55.7 & 81.2 & 75.7 &  61.9&  83.8 & 71.2\\
     SHOT\cite{SHOT} & $\sim$  & 57.1 & 78.1 & 81.5 & 68.0 & 78.2 & 78.1 & 67.4 & 54.9 & 82.2 & 73.3 & 58.8 & 84.3 & 71.8 \\
     ATDOC-NA\cite{atdoc} & $\sim$  & 58.3 & 78.8 & 82.3 & 69.4 & 78.2&  78.2&  67.1 & 56.0 & 82.7&  72.0 & 58.2 & 85.5 & 72.2\\
 \midrule
 \multicolumn{15}{c}{\textbf{DeiT} backbone} \\
 \midrule
  DeiT-S\cite{DeiT} & 21.8  & 54.4 & 73.8 & 79.9 & 68.6 & 72.6 & 75.1 & 63.6 & 50.2 & 80.0 & 73.6 & 55.2 & 82.2 & 69.1\\ 
  % \textcolor{blue}{DeiT-B}\cite{DeiT} & 60.2 & 78.3 & 82.7 & 73.3 & 77.3 & 80.3 & 69.6 & 54.9 & 82.3 & 77.3 & 59.9 & 85.2 & 73.4\\
  CDTrans-S\cite{cdtrans} & $\sim$ & 60.6 & 79.5 & 82.4 & 75.6 & 81.0 & 82.3 & 72.5 & 56.7 & 84.4 & 77.0 & 59.1 & 85.5 & 74.7 \\
  % \textcolor{blue}{CDTrans-B}\cite{cdtrans} & 68.8 & 85.0 & 86.9 & 81.5 & 87.1 & 87.3 & 79.6 & 63.3 & 88.2 & 82.0 & 66.0 & 90.6 & 80.5 \\
  WinTR-S (Ours) & $\sim$ & \textbf{65.3} & \textbf{84.1} & \textbf{85.0}  & \textbf{76.8} & \textbf{84.5} & \textbf{84.4} & \textbf{73.4}  & \textbf{60.0} & \textbf{85.7}  & \textbf{77.2} & \textbf{63.1}  & \textbf{86.8} & \textbf{77.2} \\
  % \textcolor{blue}{Ours-B} & 68.5 & 87.3 & 87.6 & 81.5 & 87.4 & 87.4 & 78.4 & 65.1 & 88.1 & 81.3 & 66.2 & 90.4 & 80.8 \\
    \bottomrule
  \end{tabular}
  }% 
  \vspace{-10pt}
\end{table*}%

\section{Experiment}
\subsection{Datasets and Setup}
We test and analyze our proposed method on three benchmark datasets in UDA, that is Office-Home~\cite{Office-Home}, VisDA-2017~\cite{VisDA2017} and DomainNet~\cite{DomainNet}. Detailed descriptions can be found in the supplementary. Note that except for special announcement, all the reported target accuracies are from the target classifier $g^t$.

% \textbf{Office-Home}~\cite{Office-Home} is a standard dataset for DA which contains four different domains: Artistic(\textbf{Ar}), Clip Art(\textbf{Cl}), Product(\textbf{Pr}) and Real-world(\textbf{Re}). Each domain consists of 65 object categories found typically in office and home scenarios.

% \textbf{VisDA-2017}~\cite{VisDA2017} is a 12-class UDA classification dataset for cross-domain tasks from synthetic(\textbf{S}) to real(\textbf{R}). Among them, the training set incorporates 152,397 synthetic images  and the validation set contains 55,388 real-world images collected from Microsoft COCO\cite{Microsoft2014}. %In our experiments, we denote training set as \textbf{S} and validation set as \textbf{R}, and construct one transfer task: \textbf{S}$\rightarrow$\textbf{R}.

% \textbf{DomainNet}~\cite{DomainNet} is currently the largest and the most challenging cross-domain benchmark. The whole dataset comprises $\sim$0.6 million images drawn from 345 categories and six diverse domains: Infograph(\textbf{inf}), Quickdraw(\textbf{qdr}), Real(\textbf{rel}), Sketch(\textbf{skt}), Clipart(\textbf{clp}), Painting(\textbf{pnt}). Thirty adaptation tasks are constructed to evaluate UDA methods, i.e., \textbf{inf$\rightarrow$qdr}, ..., \textbf{pnt$\rightarrow$clp}.
% %In particular, there are significantly different visual patterns across domains, which can be real-world images or abstract creations.
 
\subsection{Implementation Details}
Our method is based on the two-token version of DeiT-Small(\textbf{S}) and DeiT-Base(\textbf{S})~\cite{DeiT} backbone pretrained on ImageNet-1k\cite{imagenet}. We set the base learning rate as 3e-3 on VisDA2017 and DomainNet and 3e-4 on Office-Home, while both randomly initialized classifiers have a 10 times larger learning rate following~\cite{CDAN}. The training process is optimized by Stochastic Gradient Descent (SGD) with momentum 0.9 and weight decay 1e-3. The batch-size is 32 for both domains.
We adopt the source pretrained model to obtain the initial target pseudo-labels. The trade-off parameters $\lambda$ and the temperature $\tau$ are set as $1.0$ and $0.1$ respectively for all datasets.
%The training process is optimized by Stochastic Gradient Descent (SGD) with the learning rate 3e-3 and the weight decay ratio 1e-3 for all the datasets above.

\subsection{Overall Results}
We adopt DeiT-S/B as the backbones of our method and compare various UDA methods in the experiments.
As most of the convolution-based methods use ResNet-50 as their backbones on Office-Home and DomainNet, we report results using DeiT-S of similar sizes on the two benchmarks for a fair comparison and put the results with DeiT-B in the supplementary.
For VisDA-2017, both results on DeiT-S/B are reported to compare with ResNet101 backbone.
% DeiT-S/B denotes the source pretrained Transformer model.

\newcommand{\tabincell}[2]{\begin{tabular}{@{}#1@{}}#2\end{tabular}}  %表格自动换行
\begin{table*}[htbp]%%\vspace{3mm}
  \centering
  \caption{Accuracy (\%) on DomainNet for unsupervised domain adaption. In each sub-table, the column-wise domains are selected as the source domain and the row-wise domains are selected as the target domain.}
  \vspace{-8pt} 
   \resizebox{\textwidth}{!}{
  \setlength{\tabcolsep}{0.5mm}{
    \begin{tabular}{|c|ccccccc|c|ccccccc||c|ccccccc|c|ccccccc|}
    % \toprule
    \hline
    {ResNet50\cite{resnet}} & {  clp}   & {  inf}   & {  pnt}   & {  qdr}   & {  rel}   & {  skt}   & Avg.  & 
    {MCD\cite{MCD}} & {  clp}   & {  inf}   & {  pnt}   & {  qdr}   & {  rel}   & {  skt}   & Avg.  & 
    {CDAN\cite{CDAN}} & {  clp}   & {  inf}   & {  pnt}   & {  qdr}   & {  rel}   & {  skt}   & Avg.  \\
    \hline
    {  clp}   & --    & 14.2  & 29.6  & 9.5   & 43.8  & 34.3  & 26.3  & {  clp}   & --    & 15.4  & 25.5  & 3.3   & 44.6  & 31.2  & 24.0  & {  clp}   & --    & 13.5  & 28.3  & 9.3   & 43.8  & 30.2  & 25.0 \\
    {  inf}   & 21.8  & --    & 23.2  & 2.3   & 40.6  & 20.8  & 21.7  & {  inf}   & 24.1  & --    & 24.0  & 1.6   & 35.2  & 19.7  & 20.9  & {  inf}   & 16.9  & --    & 21.4  & 1.9   & 36.3  & 21.3  & 20.0 \\
    {  pnt}   & 24.1  & 15.0  & --    & 4.6   & 45.0  & 29.0  & 23.5  & {  pnt}   & 31.1  & 14.8  & --    & 1.7   & 48.1  & 22.8  & 23.7  & {  pnt}   & 29.6  & 14.4  & --    & 4.1   & 45.2  & 27.4  & 24.2 \\
    {  qdr}   & 12.2   & 1.5  & 4.9   & --    & 5.6   & 5.7   & 6.0   & {  qdr}   & 8.5   & 2.1   & 4.6   & --    & 7.9   & 7.1   & 6.0   & {  qdr}   & 11.8  & 1.2   & 4.0   & --    & 9.4   & 9.5   & 7.2  \\
    {  rel}   & 32.1  & 17.0  & 36.7  & 3.6   & --    & 26.2  & 23.1  & {  rel}   & 39.4  & 17.8  & 41.2  & 1.5   & --    & 25.2  & 25.0  & {  rel}   & 36.4  & 18.3  & 40.9  & 3.4   & --    & 24.6  & 24.7 \\
    {  skt}   & 30.4  & 11.3  & 27.8  & 3.4   & 32.9  & --    & 21.2  & {  skt}   & 37.3  & 12.6  & 27.2  & 4.1   & 34.5  & --    & 23.1  & {  skt}   & 38.2  & 14.7  & 33.9  & 7.0   & 36.6  & --    & 26.1 \\
    Avg.      & 24.1  & 11.8  & 24.4  & 4.7   & 33.6  & 23.2  & 20.3  & Avg.      & 28.1  & 12.5  & 24.5  & 2.4   & 34.1  & 21.2  & 20.5  & Avg.      & 27.0  & 12.4  & 25.7  & 5.1   & 34.3  & 22.6  & 21.2 \\
    \hline
    \hline
    BNM\cite{BNM} & {  clp}   & {  inf}   & {  pnt}   & {  qdr}   & {  rel}   & {  skt}   & Avg.  & 
    BCDM\cite{BCDM} & {  clp}   & {  inf}   & {  pnt}   & {  qdr}   & {  rel}   & {  skt}   & Avg.  &
    SCDA\cite{scda} & {  clp}   & {  inf}   & {  pnt}   & {  qdr}   & {  rel}   & {  skt}   & Avg. \\
    \hline
    {  clp}   & --    & 12.1  & 33.1  & 6.2  & 50.8  & 40.2  & 28.5  & {  clp}   & --    & 17.2  & 35.2  & 10.6  & 50.1  & 40.0  & 30.6 & {  clp}   & --     & 18.6  & 39.3  & 5.1   & 55.0  & 44.1  & 32.4 \\
    {  inf}   & 26.6  & --    & 28.5  & 2.4  & 38.5  & 18.1  & 22.8  & {  inf}   & 29.3  & --    & 29.4  & 3.8   & 41.3  & 25.0  & 25.8 & {  inf}   & 29.6   & --    & 34.0  & 1.4   & 46.3  & 25.4  & 27.3 \\
    {  pnt}   & 39.9  & 12.2  & --    & 3.4  & 54.5  & 36.2  & 29.2  & {  pnt}   & 39.2  & 17.7  & --    & 4.8   & 51.2  & 34.9  & 29.6 & {  pnt}   & 44.1   & 19.0  & --    & 2.6   & 56.2  & 42.0  & 32.8 \\
    {  qdr}   & 17.8  & 1.0   & 3.6   & --   & 9.2   & 8.3   & 8.0   & {  qdr}   & 19.4  & 2.6   & 7.2   & --    & 13.6  & 12.8  & 11.1 & {  qdr}   & 30.0   & 4.9   & 15.0  & --    & 25.4  & 19.8  & 19.0 \\
    {  rel}   & 48.6  & 13.2  & 49.7  & 3.6  & --    & 33.9  & 29.8  & {  rel}   & 48.2  & 21.5  & 48.3  & 5.4   & --    & 36.7  & 32.0 & {  rel}   & 54.0   & 22.5  & 51.9  & 2.3   & --    & 42.5  & 34.6 \\
    {  skt}   & 54.9  & 12.8  & 42.3  & 5.4  & 51.3  & --    & 33.3  & {  skt}   & 50.6  & 17.3  & 41.9  & 10.6  & 49.0  & --    & 33.9 & {  skt}   & 55.6   & 18.5  & 44.7  & 6.4   & 53.2    & --  & 35.7 \\
    Avg.      & 37.6  & 10.3  & 31.4  & 4.2  & 40.9  & 27.3  & 25.3  & Avg.      & 37.3  & 15.3  & 32.4  & 7.0   & 41.0  & 29.9  & 27.2 & Avg.      & 37.6   & 14.6  & 31.5  & 14.8  & 43.3  & 28.8  & 28.4 \\
    \hline
    \hline
    DeiT-S\cite{DeiT} & {  clp}   & {  inf}   & {  pnt}   & {  qdr}   & {  rel}   & {  skt}   & Avg.  &
    CDTrans-S\cite{cdtrans} & {  clp}   & {  inf}   & {  pnt}   & {  qdr}   & {  rel}   & {  skt}   & Avg.  & 
    WinTR-S & {  clp}   & {  inf}   & {  pnt}   & {  qdr}   & {  rel}   & {  skt}   & Avg. \\
    \hline
    {  clp}  &  --   & 20.8  & 43.2  & 14.3  & 58.8  & 46.4  & 36.7  &  {  clp}  & --    & 24.2  & 47.0  & 22.3  & 64.3  & 50.6  & 41.7  &  {  clp}  & --    & 19.9  & 53.2  & 28.6  & 70.5  & 51.6  & 44.8 \\
    {  inf}  & 35.2  &  --   & 36.7  & 4.7   & 50.4  & 30.0  & 31.4  &  {  inf}  & 45.3  & --    & 45.3  & 6.6   & 62.8  & 38.3  & 39.7  &  {  inf}  & 62.3  &  --   & 53.4  & 17.3   & 72.2  & 48.9  & 50.8   \\
    {  pnt}  & 44.7  & 20.0  &  --   & 4.5   & 59.0  & 38.1  & 33.3  &  {  pnt}  & 53.6  & 20.4  & --    & 10.6  & 63.9  & 42.4  & 38.2  &  {  pnt}  & 62.0  & 16.5  & --    & 18.0  & 71.3  & 50.1  & 43.6 \\
    {  qdr}  & 23.2  & 3.3   & 10.1  &  --   & 17.0  & 14.5  & 13.6  &  {  qdr}  & 2.8   & 0.2   & 0.6   & --    & 0.7   & 4.2   & 1.7   &  {  qdr}  & 45.1   & 4.9   & 26.1   &  --   & 17.5  & 23.5  & 23.4   \\
    {  rel}  & 48.3  & 21.3  & 50.4  & 7.0   &  --   & 37.0  & 32.8  &  {  rel}  & 47.1  & 17.9  & 45.0  & 7.9   & --    & 31.7  & 29.9  &  {  rel}  & 63.1  & 21.0  & 55.9  & 17.6  & --    & 48.8  & 41.3 \\
    {  skt}  & 54.3  & 16.5  & 41.1  & 15.3  & 53.8  &  --   & 36.2  &  {  skt}  & 61.0  & 19.3  & 46.8  & 22.8  & 59.2  & --    & 41.8  &  {  skt}  & 65.3  & 18.4  & 54.1  & 25.8  & 70.1  & --    & 46.7   \\
    Avg.     & 41.1  & 16.4  & 36.3  & 9.2   & 47.8  & 33.2  & 30.7  &  Avg.     & 42.0  & 16.4  & 36.9  & 14.0  & 50.2  & 33.4  & 32.2  &  Avg.     & 59.6 &16.1 &48.5 &21.5 &60.3 &44.6 & \textbf{41.8} \\
    \hline
    \end{tabular}%
    \label{tab:domainnet} }}%
\end{table*}%

% Table generated by Excel2LaTeX from sheet 'Sheet7'
\begin{table*}[htbp]
  \centering
  \caption{Ablation Study of Our Method on Office-Home.}
  \label{tab:ablation}%
  \vspace{-8pt} 
  \Huge
  \resizebox{0.85\textwidth}{!}{%
  \begin{tabular}{lccccccccccccc}
  \toprule
  Method & Ar$\rightarrow$Cl & Ar$\rightarrow$Pr & Ar$\rightarrow$Re & Cl$\rightarrow$Ar & Cl$\rightarrow$Pr & Cl$\rightarrow$Re & Pr$\rightarrow$Ar & Pr$\rightarrow$Cl &Pr$\rightarrow$Re & Re$\rightarrow$Ar & Re$\rightarrow$Cl & Re$\rightarrow$Pr & Avg. \\
  \midrule
    WinTR-S w/o Mask $M$ & 64.5  & 82.9  & 84.0  & 75.5  & 83.8  & 82.9  & 71.7  & 58.8  & 84.5  & 77.0  & 63.5 & 86.7  & 76.3  \\
    WinTR-S w/o $\mathcal{L}^{con}_s$ & 63.5 &82.5 &84.2 &74.7 &83.4 &83.0 &70.3 &57.5 &84.2 &76.7 &62.0 &\textbf{87.2} &75.8  \\
    WinTR-S w/o $\mathcal{L}^{con}_t$ & 64.6 &82.4 &84.3 &75.4 &84.4 &83.1 &71.5 &58.4 &84.7 &\textbf{77.3} & \textbf{63.7} &86.9 &76.4  \\
    WinTR-S w/o $\mathcal{L}^{con}_s + \mathcal{L}^{con}_t$ & 61.0 &81.8 &83.1 &73.9 &81.8 &81.5 &69.6 &56.8 &83.2 &75.6 &61.3 &86.0 &74.6 \\
    WinTR-S & \textbf{65.3} &\textbf{84.1} &\textbf{85.0} &\textbf{76.8} &\textbf{84.5} &\textbf{84.4} &\textbf{73.4} &\textbf{60.0} &\textbf{85.7} &77.2 &63.1 &86.8 &\textbf{77.2}  \\
    \bottomrule
  \end{tabular}
  }% 
  \vspace{-10pt}
\end{table*}%

\textbf{The results on VisDA-2017 and Office-Home} are reported in Table~\ref{tab:visda},\ref{tab:home}.
% Although DeiT-S is $11.5\%$ higher than ResNet-101 which shows superiority of Transformers on UDA, its performance is still way below convolution-based UDA method.
First, we notice that the source model DeiT-S surpasses ResNet-101 by $14.3\%$ on VisDA-2017 and ResNet-50 by $9.7\%$ on Office-Home, which shows the superiority of vision Transformers on UDA.
Nevertheless, its performance is still way below the convolution-based UDA method, making it necessary to design suitable algorithms to make full use of Transformers.
Second, WinTR-B achieves a class average accuracy of $90.1\%$ which outperforms all the baseline methods on VisDA-2017, and WinTR-S surpasses ATDOC-NA by an average of $5.0\%$ on Office-Home. The significant accuracy boost validates the effectiveness of our method.

% To our best knowledge, our method is the first to exceed $90\%$ accuracy on the VisDA dataset.
% We can observe that our proposed method achieves \textcolor{red}{$90.4\%$} accuracy
% Compared with convolution-based method, WinTR has no obvious defects that still has $60\%$ performance on the hard transfer truck class.
% % Our method also significantly outperforms CDTrans by $13.9\%$ on the car class.
% Moreover, our method has reached $88.4\%$ based on DeiT-S which has similar size with ResNet-50.

% \textbf{The results on Office-Home} are reported in Table~\ref{tab:home}.
% Since pseudo-labels will produce a large number of miss-matches in the presence of large number of categories, it is possible that the miss-labeled datas are harmful to the re-selection of pseudo-labels. 
% Our method arrived $77.4\%$ and $80.8\%$ by getting rid of the noise during difficult transfer tasks, e.g. \textbf{Ar$\rightarrow$Cl}, \textbf{Pr$\rightarrow$Cl}, which generated by larger domain shift.
% It is desirable that the source token of our method never fit the target domain, which eliminate the impact of noise by exploiting the involuted structural information of target datas.

% Table generated by Excel2LaTeX from sheet 'Sheet7'
\begin{table*}[htbp]
  \centering
  \Huge
  \caption{Comparison among different implementations of our proposed framework.}
  \label{tab:refine}%
  \vspace{-8pt}
  \resizebox{0.84\textwidth}{!}
  {%
  \begin{tabular}{lccccccccccccc}
  \toprule
  Method & Ar$\rightarrow$Cl & Ar$\rightarrow$Pr & Ar$\rightarrow$Re & Cl$\rightarrow$Ar & Cl$\rightarrow$Pr & Cl$\rightarrow$Re & Pr$\rightarrow$Ar & Pr$\rightarrow$Cl &Pr$\rightarrow$Re & Re$\rightarrow$Ar & Re$\rightarrow$Cl & Re$\rightarrow$Pr & Avg. \\
  \midrule
  \multicolumn{14}{c}{Label Refinement Methods} \\
  \midrule
    w/ KNN~\cite{atdoc}   & 61.3 & 81.9 & 83.0 & 71.9 &77.6 &80.2 &70.1 &55.6 &84.2 &73.9 &59.0 &84.7 &73.6  \\
    w/ K-means~\cite{SHOT} (Ours) & 65.3 &84.1 &85.0 &76.8 &84.5 &84.4 &73.4 &60.0 &85.7 &77.2 &63.1 &86.8 &\textbf{77.2}  \\
  \midrule
    \multicolumn{14}{c}{Knowledge Transfer Methods} \\
    \midrule
    w/ MDD~\cite{MMD}   & 63.7  & 83.8 & 83.9    & 75.0 & 84.2 & 83.3& 72.5 & 57.9  & 84.4 & 76.1  & 62.9 & 86.5 & 76.2 \\
    w/ MSTN~\cite{MSTN}   & 64.2  & 83.6 & 84.0    & 75.2 & 84.3 & 83.0 & 73.5 & 59.2  & 84.2 & 76.5  & 62.8 & 86.0  & 76.4 \\
    w/ Contrastive~\cite{supCon} (Ours) & 65.3 &84.1 &85.0 &76.8 &84.5 &84.4 &73.4 &60.0 &85.7 &77.2 &63.1 &86.8 &\textbf{77.2}  \\
    \bottomrule
  \end{tabular}
  }% 
\end{table*}%

\textbf{The results on DomainNet} is reported in Table~\ref{tab:domainnet}, where strong results prove our effectiveness when meeting more challenging transfer tasks and unseen target test data. 
% and we make two pairwise comparison for networks having similar amount of parameters: ResNet50 vs. DeiT-Small, ResNet152 vs. DeiT-Base (in appendix).
% The row of the table corresponds the source domain in the tasks, while the columns represent target domains.
Clearly, we observe that WinTR-S obtains a significant $13.4\%$ average accuracy boost over SCDA and a $9.6\%$ increase over CDTrans. 
Moreover, our method achieves satisfying results when adapted to \textbf{Real} images, achieving an average of $60.3\%$ accuracy. We hope these promising results on DomainNet would encourage future works to further push the limits of UDA methods on harder benchmarks.
% Moreover, note that on DomainNet we conduct contrastive learning and its reulst prevails other methods in the mean value of each column.
% It shows that two tokens encourage the network learning the knowledg that belongs solely to the each domain.
% More results shown in Table\ref{tab:ablation} support our conclusion.

\subsection{Insight Analysis}\label{sec:analysis}
In this section, we carry out experiments to fully investigate the influence of each component in our win-win Transformer framework. All the analytical experiments are based on the DeiT-S backbone.

\begin{figure*}[tb]
  \centering
  \includegraphics[width=0.98\textwidth]{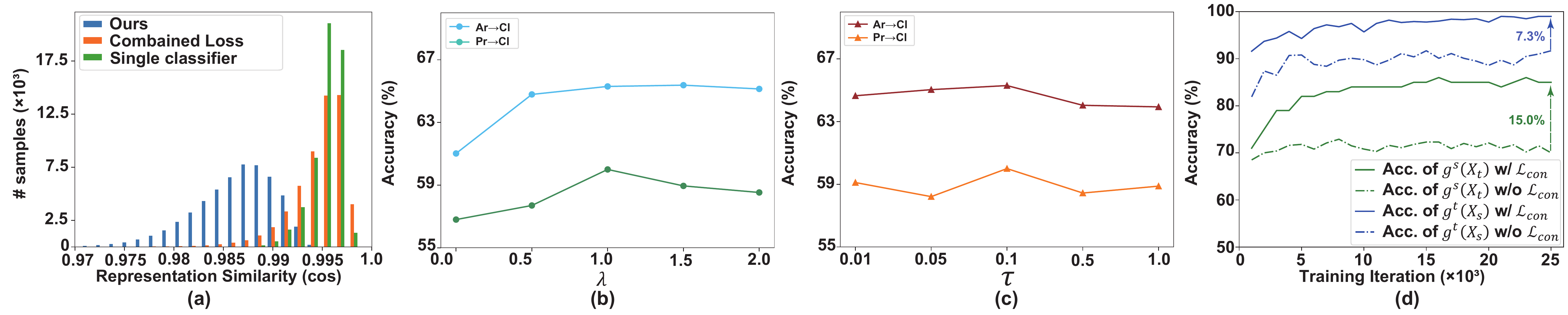}
  \vspace{-5pt}
  \caption{Analysis experiments: (a) Token similarity for all target samples in VisDA-2017, (b) and (c) Parameter sensitivity analysis on Office-Home, (d) Improvements on generalization brought by contrastive-based cross-domain knowledge transfer.}
  \label{Fig_analysis}
\end{figure*}

\textbf{Domain-specific Knowledge Learning.}
To show that the two tokens in our method preserve more domain-specific information than traditional framework does, we examine the cosine similarity between source-oriented representation $\f^s_{tj}$ and target-oriented representation $\f^t_{tj}$ for each target sample. We compare our method with two variants: a) optimizing both classifiers with the same loss $\mathcal{L}_s+\mathcal{L}_t$, b) replacing the two classifiers by one common classifier trained by $\mathcal{L}_s+\mathcal{L}_t$. The results of the similarity score for all target samples are shown in Fig.~\ref{Fig_analysis}(a). We find that more target samples obtain divergent representations on the two tokens under our proposed separate training framework, indicating that they learn different perspectives. 

\textbf{Ablation Study.}
To examine the influence of components on the final performance, we respectively remove the domain-wise token mask and contrastive losses.
The results are listed in Table~\ref{tab:ablation}. It can be found that each component contributes to our method, especially on harder transfer tasks that need to exploit more knowledge of both in-domain and cross-domain. We also note that the mask plays an important role by preventing the two tokens from interfering with each other.
% It can be found that the more difficult transfer task is, the less knowledge of target domain the network has.
% Contrastive learning let the target token self-cultivation without the impact of source to overcome obstacles.

\textbf{Sensitivity Analysis.}
To show that our method is robust to different choices on hyperparameter $\lambda$ and $\tau$, we vary their values on two randomly selected Office-Home tasks and show the results in Fig.~\ref{Fig_analysis}(b) and (c). %需确认
Expect for setting $\lambda=0$ where the contrastive loss is aborted, our method maintains stable accuracies in spite of the parameter change. 
% Actually, we find $\lambda = 1$ achieves satisfying results on all datasets with no need for special annealing.

\textbf{Alternatives for Label Refinement and Knowledge Transfer Methods.}
The proposed WinTR could be a general framework with multiple options. We conduct experiments on different alternatives for the label refinement and knowledge transfer in our method. 

The alternatives (including ours) for label refinement that utilize structural information in the feature space are: 
\textit{K-Nearest Neighbor (KNN)}~\cite{atdoc}, where the label of a target sample is determined by its nearest $K$ samples in the feature space. Here we set the hyperparameter $K=5$.
\textit{K-means}~\cite{SHOT}, which is adopted and described in our method.

The alternatives for knowledge transfer that aligns the source and target features for each domain-oriented representation are:
\textit{Maximum Mean Discrepancy (MMD)}~\cite{MMD}, \textit{MSTN}~\cite{MSTN} that aligns the source feature center with the target feature center of the same class and \textit{Contrastive} inspired by~\cite{supCon}, which is described in our method.
Note that we keep the stop-grad operation on $\f^s_{s}$ and $\f^t_{t}$ for each knowledge transfer method to maintain the integrity of the two domain-oriented representations.

% \textcolor{blue}{The results in Table~\ref{tab:refine} show that different region based label refinement can obtain the less noise of pseudo label from source token. Although, the source classifier is never trained by target data, label refine could employ features to improve the accuracy of source classifier.}

% \begin{figure}[htb]
%   \centering
%   \includegraphics[width=0.48\textwidth]{figures/Fig_contrastive.pdf}
%   \vspace{-15pt}
%   \caption{Improvements on generalization brought by contrastive-based cross-domain knowledge transfer. }
%   \label{Fig_contrastive}
% \end{figure}

\textbf{Contrastive-based Knowledge Transfer Improves Generalization.}
Here we further investigate the effect of the contrastive-based knowledge transfer. We test target data on source-specific classifier, (i.e. $\boldsymbol{g}^s(X_t)$) and vise versa. Note that in this experiment, the test data comes from the other domain and is never trained on the classifier being tested.
As shown in Fig.~\ref{Fig_analysis}(d), after applying both contrastive losses (denote as w/ $\mathcal{L}_{con}$), the prediction accuracies on unseen data from the other domain improve significantly, verifying that the cross-domain knowledge transfer learns better feature embeddings than only conducting in-domain knowledge exploration.

\begin{table}[htbp]
  \centering
  \caption{Comparison between using source-oriented and target-oriented representation for label refinement.}
  \label{tab:target_refine}%
  \vspace{-8pt}
  \resizebox{0.49\textwidth}{!}{%
  \begin{tabular}{lcccc|c}
  \toprule
  Variants & Cl$\rightarrow$Ar & Pr$\rightarrow$Cl & Re$\rightarrow$Cl & Home Avg. & Visda-2017 \\
  \midrule
  target-oriented &75.3 &56.8 &60.7 &76.2 & 86.4\\
  source-oriented &76.8 &60.0 &63.1 &77.2 & 88.8\\
    \bottomrule
  \end{tabular}
  }% 
\end{table}%

\textbf{Which Representation for Label Refinement?}
To prove that the label refinement process achieves better results when utilizing source-oriented representation instead of target-oriented ones, we compare the two variants on several tasks in Office-Home and VisDA-2017. Specifically, we employ the same label refinement method on $\f^s_{t}$ and $\f^t_{t}$ respectively and use the obtained pseudo-labels for model training. The results are listed in Table.~\ref{tab:target_refine}, which indicates that source-oriented representation is more suitable, especially for harder transfer tasks like Pr$\rightarrow$Cl when pseudo-labels before refinement are noisier.

\section{Conclusion}
We propose a win-win Transformer framework for UDA. Different from the classical UDA paradigm that learns a domain-invariant representation and reuses the source classifier to predict target, we propose to separately train two mappings and two domain-specific classifiers via [src] and [tgt] token. The new framework enables a simultaneous exploitation of domain-specific and invariant knowledge without interfering with each other. We propose two mechanisms to conduct cross-domain knowledge transfer between the two tokens while training the two classifiers with different objectives. Extensive experiments on three benchmarks prove the effectiveness of our method.
% Extensive experiments on three benchmarks demonstrate that our method substantially outperforms baseline methods using convolution-based or attention-based backbones.

% %%%%%%%%% REFERENCES
% \clearpage
\section{Supplementary Material}
\begin{table*}[htbp]
  \centering
  \caption{Complete table of accuracy (\%) results on DomainNet for unsupervised domain adaption. In each sub-table, the column-wise domains are selected as the source domain and the row-wise domains are selected as the target domain. Methods from the first two rows are based on ResNet-50 backbone, while the methods from the last two rows are based on DeiT-Small/Base architectures.}
  % \vspace{-8pt} 
   \resizebox{\textwidth}{!}{
  \setlength{\tabcolsep}{0.5mm}{
    \begin{tabular}{|c|ccccccc|c|ccccccc||c|ccccccc|c|ccccccc|}
    \hline
    {ResNet50\cite{resnet}} & {  clp}   & {  inf}   & {  pnt}   & {  qdr}   & {  rel}   & {  skt}   & Avg.  & 
    {MCD\cite{MCD}} & {  clp}   & {  inf}   & {  pnt}   & {  qdr}   & {  rel}   & {  skt}   & Avg.  & 
    {CDAN\cite{CDAN}} & {  clp}   & {  inf}   & {  pnt}   & {  qdr}   & {  rel}   & {  skt}   & Avg.  \\
    \hline
    {  clp}   & --    & 14.2  & 29.6  & 9.5   & 43.8  & 34.3  & 26.3  & {  clp}   & --    & 15.4  & 25.5  & 3.3   & 44.6  & 31.2  & 24.0  & {  clp}   & --    & 13.5  & 28.3  & 9.3   & 43.8  & 30.2  & 25.0 \\
    {  inf}   & 21.8  & --    & 23.2  & 2.3   & 40.6  & 20.8  & 21.7  & {  inf}   & 24.1  & --    & 24.0  & 1.6   & 35.2  & 19.7  & 20.9  & {  inf}   & 16.9  & --    & 21.4  & 1.9   & 36.3  & 21.3  & 20.0 \\
    {  pnt}   & 24.1  & 15.0  & --    & 4.6   & 45.0  & 29.0  & 23.5  & {  pnt}   & 31.1  & 14.8  & --    & 1.7   & 48.1  & 22.8  & 23.7  & {  pnt}   & 29.6  & 14.4  & --    & 4.1   & 45.2  & 27.4  & 24.2 \\
    {  qdr}   & 12.2   & 1.5  & 4.9   & --    & 5.6   & 5.7   & 6.0   & {  qdr}   & 8.5   & 2.1   & 4.6   & --    & 7.9   & 7.1   & 6.0   & {  qdr}   & 11.8  & 1.2   & 4.0   & --    & 9.4   & 9.5   & 7.2  \\
    {  rel}   & 32.1  & 17.0  & 36.7  & 3.6   & --    & 26.2  & 23.1  & {  rel}   & 39.4  & 17.8  & 41.2  & 1.5   & --    & 25.2  & 25.0  & {  rel}   & 36.4  & 18.3  & 40.9  & 3.4   & --    & 24.6  & 24.7 \\
    {  skt}   & 30.4  & 11.3  & 27.8  & 3.4   & 32.9  & --    & 21.2  & {  skt}   & 37.3  & 12.6  & 27.2  & 4.1   & 34.5  & --    & 23.1  & {  skt}   & 38.2  & 14.7  & 33.9  & 7.0   & 36.6  & --    & 26.1 \\
    Avg.      & 24.1  & 11.8  & 24.4  & 4.7   & 33.6  & 23.2  & 20.3  & Avg.      & 28.1  & 12.5  & 24.5  & 2.4   & 34.1  & 21.2  & 20.5  & Avg.      & 27.0  & 12.4  & 25.7  & 5.1   & 34.3  & 22.6  & 21.2 \\
    \hline
    \hline
    BNM\cite{BNM} & {  clp}   & {  inf}   & {  pnt}   & {  qdr}   & {  rel}   & {  skt}   & Avg.  & 
    BCDM\cite{BCDM} & {  clp}   & {  inf}   & {  pnt}   & {  qdr}   & {  rel}   & {  skt}   & Avg.  &
    SCDA\cite{scda} & {  clp}   & {  inf}   & {  pnt}   & {  qdr}   & {  rel}   & {  skt}   & Avg. \\
    \hline
    {  clp}   & --    & 12.1  & 33.1  & 6.2  & 50.8  & 40.2  & 28.5  & {  clp}   & --    & 17.2  & 35.2  & 10.6  & 50.1  & 40.0  & 30.6 & {  clp}   & --     & 18.6  & 39.3  & 5.1   & 55.0  & 44.1  & 32.4 \\
    {  inf}   & 26.6  & --    & 28.5  & 2.4  & 38.5  & 18.1  & 22.8  & {  inf}   & 29.3  & --    & 29.4  & 3.8   & 41.3  & 25.0  & 25.8 & {  inf}   & 29.6   & --    & 34.0  & 1.4   & 46.3  & 25.4  & 27.3 \\
    {  pnt}   & 39.9  & 12.2  & --    & 3.4  & 54.5  & 36.2  & 29.2  & {  pnt}   & 39.2  & 17.7  & --    & 4.8   & 51.2  & 34.9  & 29.6 & {  pnt}   & 44.1   & 19.0  & --    & 2.6   & 56.2  & 42.0  & 32.8 \\
    {  qdr}   & 17.8  & 1.0   & 3.6   & --   & 9.2   & 8.3   & 8.0   & {  qdr}   & 19.4  & 2.6   & 7.2   & --    & 13.6  & 12.8  & 11.1 & {  qdr}   & 30.0   & 4.9   & 15.0  & --    & 25.4  & 19.8  & 19.0 \\
    {  rel}   & 48.6  & 13.2  & 49.7  & 3.6  & --    & 33.9  & 29.8  & {  rel}   & 48.2  & 21.5  & 48.3  & 5.4   & --    & 36.7  & 32.0 & {  rel}   & 54.0   & 22.5  & 51.9  & 2.3   & --    & 42.5  & 34.6 \\
    {  skt}   & 54.9  & 12.8  & 42.3  & 5.4  & 51.3  & --    & 33.3  & {  skt}   & 50.6  & 17.3  & 41.9  & 10.6  & 49.0  & --    & 33.9 & {  skt}   & 55.6   & 18.5  & 44.7  & 6.4   & 53.2    & --  & 35.7 \\
    Avg.      & 37.6  & 10.3  & 31.4  & 4.2  & 40.9  & 27.3  & 25.3  & Avg.      & 37.3  & 15.3  & 32.4  & 7.0   & 41.0  & 29.9  & 27.2 & Avg.      & 37.6   & 14.6  & 31.5  & 14.8  & 43.3  & 28.8  & 28.4 \\
    \hline
    \hline
    DeiT-S\cite{DeiT} & {  clp}   & {  inf}   & {  pnt}   & {  qdr}   & {  rel}   & {  skt}   & Avg.  &
    CDTrans-S\cite{cdtrans} & {  clp}   & {  inf}   & {  pnt}   & {  qdr}   & {  rel}   & {  skt}   & Avg.  & 
    WinTR-S & {  clp}   & {  inf}   & {  pnt}   & {  qdr}   & {  rel}   & {  skt}   & Avg. \\
    \hline
    {  clp}  &  --   & 20.8  & 43.2  & 14.3  & 58.8  & 46.4  & 36.7  &  {  clp}  & --    & 24.2  & 47.0  & 22.3  & 64.3  & 50.6  & 41.7  &  {  clp}  & --    & 19.9  & 53.2  & 28.6  & 70.5  & 51.6  & 44.8 \\
    {  inf}  & 35.2  &  --   & 36.7  & 4.7   & 50.4  & 30.0  & 31.4  &  {  inf}  & 45.3  & --    & 45.3  & 6.6   & 62.8  & 38.3  & 39.7  &  {  inf}  & 62.3  &  --   & 53.4  & 17.3   & 72.2  & 48.9  & 50.8   \\
    {  pnt}  & 44.7  & 20.0  &  --   & 4.5   & 59.0  & 38.1  & 33.3  &  {  pnt}  & 53.6  & 20.4  & --    & 10.6  & 63.9  & 42.4  & 38.2  &  {  pnt}  & 62.0  & 16.5  & --    & 18.0  & 71.3  & 50.1  & 43.6 \\
    {  qdr}  & 23.2  & 3.3   & 10.1  &  --   & 17.0  & 14.5  & 13.6  &  {  qdr}  & 2.8   & 0.2   & 0.6   & --    & 0.7   & 4.2   & 1.7   &  {  qdr}  & 45.1   & 4.9   & 26.1   &  --   & 17.5  & 23.5  & 23.4   \\
    {  rel}  & 48.3  & 21.3  & 50.4  & 7.0   &  --   & 37.0  & 32.8  &  {  rel}  & 47.1  & 17.9  & 45.0  & 7.9   & --    & 31.7  & 29.9  &  {  rel}  & 63.1  & 21.0  & 55.9  & 17.6  & --    & 48.8  & 41.3 \\
    {  skt}  & 54.3  & 16.5  & 41.1  & 15.3  & 53.8  &  --   & 36.2  &  {  skt}  & 61.0  & 19.3  & 46.8  & 22.8  & 59.2  & --    & 41.8  &  {  skt}  & 65.3  & 18.4  & 54.1  & 25.8  & 70.1  & --    & 46.7   \\
    Avg.     & 41.1  & 16.4  & 36.3  & 9.2   & 47.8  & 33.2  & 30.7  &  Avg.     & 42.0  & 16.4  & 36.9  & 14.0  & 50.2  & 33.4  & 32.2  &  Avg.     & 59.6 &16.1 &48.5 &21.5 &60.3 &44.6 & \textbf{41.8} \\
    \hline
    DeiT-B\cite{DeiT} & {  clp}   & {  inf}   & {  pnt}   & {  qdr}   & {  rel}   & {  skt}   & Avg.  &
    CDTrans-B\cite{cdtrans} & {  clp}   & {  inf}   & {  pnt}   & {  qdr}   & {  rel}   & {  skt}   & Avg.  & 
    WinTR-B & {  clp}   & {  inf}   & {  pnt}   & {  qdr}   & {  rel}   & {  skt}   & Avg. \\
    \hline
    {  clp}  &   --    & 25.2  & 46.2  & 13.0  & 62.3  & 48.8  & 39.1 & {  clp}  & --    & 27.9  & 57.6  & 27.9  & 73.0  & 58.8  & 49.0  &  {  clp}  & --    & 21.6  & 56.3  & 31.6  & 72.8  & 57.3  & 47.9 \\
    {  inf}  &   46.4  & --    & 45.2  & 5.1   & 62.3  & 37.5  & 39.3 & {  inf}  & 58.6  & --    & 53.4  & 9.6   & 71.1  & 47.6  & 48.1  &  {  inf}  & 66.8 &  --   & 56.8  & 19.9   & 73.9  & 53.7  & 54.2   \\
    {  pnt}  &   48.1  & 22.1  & --    & 4.4   & 62.5  & 41.8  & 35.8 & {  pnt}  & 60.7  & 24.0  & --    & 13.0  & 69.8  & 49.6  & 43.4  &  {  pnt}  & 69.2  & 21.5  & --    & 21.3  & 74.4  & 55.6  & 48.4 \\
    {  qdr}  &   28.2  & 5.2   & 14.4  & --    & 21.9  & 17.7  & 17.5 & {  qdr}  & 2.9   & 0.4   & 0.3   & --    & 0.7   & 4.7   & 1.8  &  {  qdr}  & 52.3  & 6.3  & 28.7  &  --   & 47.3  & 38.7  & 34.7   \\
    {  rel}  &   53.2  & 24.1  & 53.5  & 7.2   & --    & 41.6  & 35.9 & {  rel}  & 49.3  & 18.7  & 47.8  & 9.4   & --    & 33.5  & 31.7  &  {  rel}  & 68.2  & 22.2  & 59.8  & 20.9  & --    & 55.1  & 45.2 \\
    {  skt}  &   58.0  & 21.8  & 46.5  & 15.7  & 58.7  & --    & 40.1 & {  skt}  & 66.8  & 23.7  & 54.6  & 27.5  & 68.0  & --    & 48.1  &  {  skt}  & 69.9  & 22.6  & 58.1  & 28.8  & 73.1  & --    &  50.5  \\
    Avg.     &   46.8  & 19.7  & 41.2  & 9.1   & 53.5  & 37.5  & 34.6 &  Avg.    & 47.7  & 18.9  & 42.7  & 17.5  & 56.5  & 38.8  & 37.0  &  Avg.     & 65.3 & 18.9 &51.9 &24.5 & 68.3& 52.1&\textbf{46.8} \\
    \hline
    \end{tabular}%
    \label{tab:domainnetbase} }}%
\end{table*}%

\subsection{Training Algorithm of WinTR}
We show in this section the overall training procedure of our proposed WinTR. As shown in Algorithm~\ref{alg:Framwork}, we first train a source model from the ImageNet-pretrained DeiT to obtain the initial target pseudo-labels for each target sample (Stage 1). Then we Reinitialize the backbone along with two new classifiers. Four losses are computed and the model is optimized. After certain rounds of training, the target pseudo-labels are updated using new features and predictions. This process continues until the model converges or the max iteration is reached (stage 2).

\begin{algorithm}[!htbp]
  \small
  \caption{\small Training Algorithm of WinTR}
  \label{alg:Framwork}
  \begin{algorithmic} [1]
    \REQUIRE
      Source domain $\{\left(\x_{si},y_{si}\right)\}_{i=1}^{n_s}$; Target domain $\{\x_{tj}\}_{j=1}^{n_t}$; Hyper-parameters $\lambda$ and $\tau$; Max iteration: $I$
    \ENSURE
      Trained model for target domain: $g^t \circ f^t$
    \begin{enumerate}
        \item[\textbf{Stage 1}] Obtain Initial Pseudo-labels for Target Domain:
    \end{enumerate}
        \STATE Initialize DeiT with parameters pretrained on ImageNet-1k;
        \STATE Train DeiT backbone in source domain using $\mathcal{L}_s$ in Eq.~(4);
        \STATE Obtain the initial target pseudo-labels $\tilde{y}_{tj}$ by Eq.~(6);
    \begin{enumerate}
        \item[\textbf{Stage 2}] Knowledge Exploration and Transfer:
    \end{enumerate}
        \STATE Reinitialize DeiT and randomly initialize two classifiers;
        \FOR{$i=1,2,\cdots,I$}
            \STATE Obtain $\f^s$ and $\f^t$ from the Transformer backbone;
            \STATE Compute $\mathcal{L}_s$ by Eq.~(4) with source classifier $g^s$ and $y_{si}$;
            \STATE Compute $\mathcal{L}_t$ by Eq.~(5) with target classifier $g^t$ and $\tilde{y}_{tj}$;
            \STATE Calculate $\mathcal{L}_s^{con}$ on $\f^s$ with stop-gradient on $\f^s_{tj}$;
            \STATE Calculate $\mathcal{L}_t^{con}$ on $\f^t$ with stop-gradient on $\f^t_{si}$;
            % \STATE Using $\textbf{g}^s_{si},\textbf{f}^s$ to calculate $L_s$ and $L_s^{con}$;
            % \STATE Using $\textbf{g}^t_{tj} , \textbf{t}^t$ to calculate $L_t$ and $L_t^{con}$;
            \STATE Accumulate the above losses and optimize the model;
            \STATE For each $\x_{tj}$, obtain new pseudo-label $\tilde{y}_{tj}$ using Eq.~(6);
        \ENDFOR
  \end{algorithmic}
\end{algorithm}

\subsection{Experimental details}
Three benchmarks in UDA that is used in our experiment are Office-Home, VisDA-2017 and DomainNet. 

\textbf{Office-Home}~\cite{Office-Home} is a standard dataset for DA which contains four different domains: Artistic(\textbf{Ar}), Clip Art(\textbf{Cl}), Product(\textbf{Pr}) and Real-world(\textbf{Re}). Each domain consists of 65 object categories found typically in office and home scenarios.

\textbf{VisDA-2017}~\cite{VisDA2017} is a 12-class UDA classification dataset for cross-domain tasks from synthetic(\textbf{S}) to real(\textbf{R}). Among them, the training set incorporates 152,397 synthetic images  and the validation set contains 55,388 real-world images collected from Microsoft COCO\cite{Microsoft2014}. %In our experiments, we denote training set as \textbf{S} and validation set as \textbf{R}, and construct one transfer task: \textbf{S}$\rightarrow$\textbf{R}.

\textbf{DomainNet}~\cite{DomainNet} is currently the largest and the most challenging cross-domain benchmark. The whole dataset comprises $\sim$0.6 million images drawn from 345 categories and six diverse domains: Infograph(\textbf{inf}), Quickdraw(\textbf{qdr}), Real(\textbf{rel}), Sketch(\textbf{skt}), Clipart(\textbf{clp}), Painting(\textbf{pnt}). Thirty adaptation tasks are constructed to evaluate UDA methods, i.e., \textbf{inf$\rightarrow$qdr}, ..., \textbf{pnt$\rightarrow$clp}.
%In particular, there are significantly different visual patterns across domains, which can be real-world images or abstract creations.

In the training procedure, we adopt similar data-augmentation techniques such as Rand-Augment~\cite{randaugment} and random erasing~\cite{random_erase} to fully utilize the limited training data as in DeiT. We find these data-augmentations very helpful for training Transformers in UDA tasks.

\subsection{Additional Experimental Results on DomainNet Dataset}
In this section, we report the complete results of our method on DomainNet, including WinTR-S and WinTR-B, and compare them to other baseline methods. The results are shown in Table~\ref{tab:domainnetbase} in the next page. We can conclude that WinTR achieves superior performance on DomainNet benchmark, reaching an average accuracy of 46.8\% using WinTR-B which is 12.2\% higher than its backbone DeiT-B. Also, WinTR-S already outperforms CDTrans-B by an average of 4.8\% as well as all the UDA methods that are based on ResNet-50 backbone. These results validate the effectiveness of our method, proving that training an individual target classifier using only target data and pseudo-labels is applicable when using a Transformer as backbone. This success might lead to the rethinking of sharing a common classifier for both source and target domain.

\newpage
{\small
\bibliographystyle{ieee_fullname}
\bibliography{Reference_CVPR2022}
}

\end{document}